\pdfoutput=1

\documentclass[11pt]{article}

\usepackage[]{acl}

\usepackage{times}
\usepackage{latexsym}
\usepackage{bm}
\usepackage{amsthm,amsmath,amssymb}
\usepackage{mathrsfs}
\usepackage{epsfig}
\usepackage{graphicx}
\usepackage{booktabs}
\usepackage{verbatim}
\usepackage{enumitem,kantlipsum}
\usepackage{tipa}
\usepackage{algorithm}
\usepackage{algorithmicx}
\usepackage{algpseudocode}

\usepackage[T1]{fontenc}

\usepackage[utf8]{inputenc}

\usepackage{microtype}

\usepackage{color} 
\definecolor{mypink2}{RGB}{219, 48, 122}
\definecolor{orange}{RGB}{255, 147, 00}
\definecolor{jrcolor}{RGB}{100, 150, 225}
\definecolor{jrcomment}{RGB}{70, 200, 150}

\algnewcommand\algorithmicinput{\textbf{Input:}}
\algnewcommand\INPUT{\item[\algorithmicinput]}
\algnewcommand\algorithmicoutput{\textbf{Output:}}
\algnewcommand\OUTPUT{\item[\algorithmicoutput]}

\title{Visualizing the Relationship Between Encoded Linguistic Information and Task Performance}

\author{Jiannan Xiang$^\heartsuit$\thanks{~~Equal contribution. Work done while J. Xiang was an intern at Tencent AI Lab.}\ , Huayang Li$^{\spadesuit*}$, Defu Lian$^\diamondsuit$, Guoping Huang$^\clubsuit$, \\ \textbf{ Taro Watanabe$^\spadesuit$, Lemao Liu$^\clubsuit$} \\
  $^\heartsuit$Carnegie Mellon University \quad $^\spadesuit$Nara Institute of Science and Technology \\ $^\diamondsuit$University of Science and Techology of China \quad $^\clubsuit$Tencent AI Lab\\
  \texttt{jiannanx@cs.cmu.edu, li.huayang.lh6@is.naist.jp} \\ \texttt{liandefu@ustc.edu.cn, donkeyhuang@tencent.com} \\ \texttt{taro@is.naist.jp, lemaoliu@gmail.com}
  }

\begin{document}
\setlength{\abovedisplayskip}{4pt}
\setlength{\belowdisplayskip}{4pt}
\maketitle
\begin{abstract}
Probing is popular to analyze whether linguistic information can be captured by a well-trained deep neural model, but it is hard to answer how the change of the encoded linguistic information will affect task performance. To this end, we study the dynamic relationship between the encoded linguistic information and task performance from the viewpoint of Pareto Optimality. Its key idea is to obtain a set of models which are Pareto-optimal in terms of both objectives. From this viewpoint, we propose a method to optimize the Pareto-optimal models by formalizing it as a multi-objective optimization problem.
We conduct experiments on two popular NLP tasks, \emph{i.e}., machine translation and language modeling, and investigate the relationship between several kinds of linguistic information and task performances. Experimental results demonstrate that the proposed method is better than a baseline method. Our empirical findings suggest that some syntactic information is helpful for NLP tasks whereas encoding more syntactic information does not necessarily lead to better performance, because the model architecture is also an important factor.
\end{abstract}

\section{Introduction}

Recent years have witnessed great success of deep neural networks for natural language processing tasks, such as language modeling~\cite{zaremba2014recurrent,merity2017regularizing} and Neural Machine Translation \citep{bahdanau2014neural,vaswani2017attention}. The excellent task performance they achieved spiked the interest in interpreting their underlying mechanism. Since linguistic knowledge is crucial in natural languages, an emerging body of literature uses \textit{probes} \citep{conneau2018you,alt2020probing,saleh2020probing,cao-etal-2021-low} to investigate whether a standard model trained towards better task performance also captures the linguistic information. From the perspective of information theory, \citet{voita2020information}  and \citet{pimentel2020information} show that probes can be used to estimate the amount of linguistic information captured by a fixed model. 


However, the above probing only extracts linguistic information from a fixed standard model, which helps little to understand the relationship between the task performance and linguistic information encoded by the model. For example, under their methodology, it is difficult to answer the following two questions. First, would adding linguistic information be beneficial for an NLP model; second, is it harmful when this linguistic information is reduced.
Therefore, it is still an open and intriguing question to reveal how task performance changes with respect to different amounts of linguistic information.


\begin{figure}[]
    \begin{center}
        \includegraphics[width=0.6\linewidth]{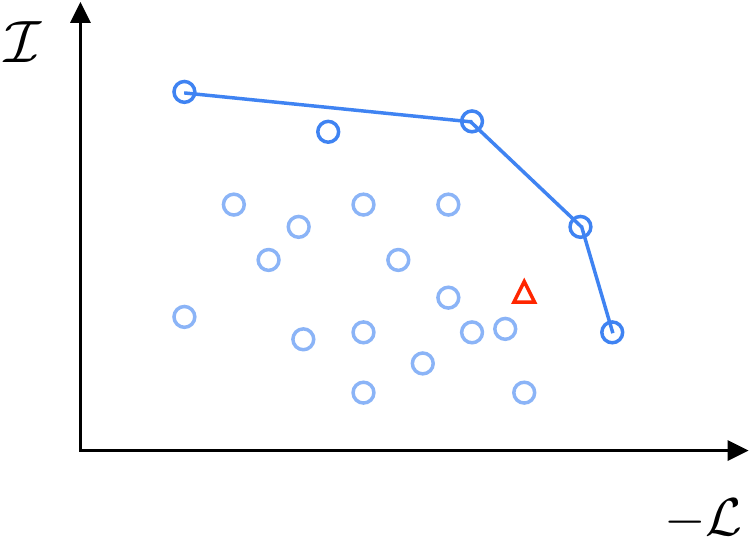}
    \end{center}
    \vspace{-2ex}
        \caption{Illustration of Pareto frontier by a toy example. Triangle ($\bigtriangleup$) corresponds to the standard checkpoint with best performance and each circle ($\bigcirc$) corresponds to a sampled checkpoint. The y-axis indicates the linguistic information $\mathcal{I}$ encoded by the model, and x-axis indicates the negative loss value $\mathcal{-L}$.\label{fig:illustration}}
        \vspace{-2ex}
\end{figure}

To this end, this paper proposes a novel viewpoint to study the relationship between task performance and the amount of linguistic information, inspired by the criterion of Pareto Optimality which is widely used in economics~\cite{greenwald1986externalities}. Our main idea is to obtain Pareto-optimal models on a test set in terms of both linguistic information and task performance and then visualize their relationship along with these optimal models. By comparing a standard model with these optimal models, it is clear to answer the question that whether adding the encoded information is helpful to improve the task performance over the standard model
, as illustrated in Figure~\ref{fig:illustration}, where the points on the line are Pareto-optimal and the red triangle denotes the standard model with best performance. 

Nevertheless, it is typically intractable to obtain the Pareto-optimal models according to both dimensions on test data. To address the challenge, we propose a principled method to approximately optimize the Pareto-optimal models on the training data which can be expected to generalise well on test sets according to statistical learning theory~\cite{vapnik1999overview}. Formally, the approach can be regarded as a multi-objective optimization problem: during the learning procedure, it optimizes two objectives, \emph{i.e}., the task performance and extracted linguistic information. In addition, we develop a computationally efficient algorithm to address the optimization problem. By inspecting the trend of those Pareto-optimal points, the relationship between task performance and linguistic information can be clearly illustrated. Back to our questions, we also consider two instances within the proposed methodology: one aims to maximize the amount of linguistic information (\emph{i.e}., adding) while the other tries to minimize it (\emph{i.e}., reducing).

We conduct experiments on two popular NLP tasks,
i.e., machine translation and language modeling,
and choose three different linguistic properties, including two syntactic properties (Part-of-Speech and dependency labels) and one phonetic property. We investigate the relationship between NMT performance and each syntactic information,  and the relationship between LM performance and phonetic information. 
For machine translation,
we use LSTM, \emph{i.e}., RNN-search \citep{bahdanau2014neural}, and Transformer \citep{vaswani2017attention} as the main model architectures, and conduct our experiments on En $\Rightarrow$ De and Zh $\Rightarrow$ En tasks. For language modeling, we employ the LSTM model and conduct experiments on the Penn Treebank dataset. 
The experimental results show that: i) syntactic information encoded by NMT models is important for MT task and reducing it leads to sharply decreased performance; ii) the standard NMT model obtained by maximum likelihood estimation (MLE) is Pareto-optimal for Transformer but it is not the case for LSTM based NMT; iii) reducing the phonetic information encoded by LM models only makes task performance drop slightly. 


In summary, our contributions are three-fold:
\begin{enumerate}[wide=0\parindent,noitemsep, topsep=0pt]
    \item We make an initial attempt to study the relationship between encoded linguistic information and task performance, \emph{i.e.}, how the change of linguistic information affects the performance of models.
    \item  We propose a new viewpoint from Pareto Optimality as well as a principled approach which is formulated as a multi-objective optimization problem, to visualize the relationship.
    \item Our experimental results show that encoding more linguistic information is not necessary to yield better task performance depending on the specific model architecture.
\end{enumerate}

\section{Related Work}

\paragraph{Probe}
With the impressive performance of Neural Network models for NLP tasks \citep{sutskever2014sequence,luong2015effective,vaswani2017attention,devlin2018bert,xu2020layoutlm}, people are becoming interested in understanding neural models~\cite{ding-etal-2017-visualizing,li-etal-2019-word,li-etal-2020-evaluating}. 
 One popular interpretation method is probe \citep{conneau2018you}, also known as auxiliary prediction \citep{adi2016fine} and diagnostic classification \citep{hupkes2018visualisation}, which aims to understand how neural models work and what information they have encoded and used. From the perspective of information theory, \citet{voita2020information} and \citet{pimentel2020information} show that probes can be used to estimate the amount of linguistic information captured by a model. However, recent research studies point out that probes fail to demonstrate whether the information is used by models. For example, \citet{hewitt2019designing} show that the probe can also achieve high accuracy in predicting randomly generated tags, which is useless for the task. And \citet{ravichander-etal-2021-probing} present that the representations encode the linguistic properties even if they are invariant and not required for the task. Instead of studying the encoded linguistic information by training a probe for fixed representations, in this work we study how the amount change of linguistic information affects the performance of NLP tasks.

\paragraph{Information Removal} Information removal is crucial in the area of transfer learning \citep{ganin2015unsupervised,tzeng2017adversarial,long2018conditional} and fairness learning \citep{xie2017controllable,elazar2018adversarial}, where people want to remove domain information or bias from learned representations. One popular method is Adversarial Learning \citep{goodfellow2014generative,ganin2015unsupervised}, which trains a classifier to predict the properties of representations, \emph{e.g.}, domain information or gender bias, while the feature extractor tries to fool the classifier. In this work, when using our method to reduce the linguistic information in the representations, we find that our multi-objective loss function is the same form as adversarial learning, which provides the theoretical guarantee for using adversarial learning to find the Pareto-optimal solutions to a multi-objective problem.

Recently, \citet{elazar2020bert} also propose to study the role of linguistic properties with the idea of information removal  \citep{ravfogel2020null}. However, the representations got by their method may not be Pareto-optimal, because it only minimizes the mutual information, but ignores the objective of task performance. On the contrary, our proposed method optimizes towards both objectives, thus our results can be used to visualize the relationship between linguistic properties and task performance.

\paragraph{Pareto Optimality} The idea of Pareto Optimality \citep{mas1995microeconomic} is an important criterion in economics, where the goal is to characterize situations where no variable can be better off without making at least one variable worse off. It has been also widely used in the area of sociology and game theory~\cite{beckman2002envy,chinchuluun2008Pareto}. In addition, in artificial intelligence \citet{martinez2020minimax} use Pareto optimality to solve group fairness problem and ~\citet{duh2012learning} proposed to optimize an MT system on multiple metrics based on the theory of Pareto optimality. In particular, \citet{pimentel-etal-2020-Pareto} propose a variant of probing on the hidden representation of deep models and they consider Pareto optimality in terms of both objectives similar to our work. Comparing with their work, one difference is the choice of objectives. Another significant difference is that they optimize probing model in a conventional fashion, and thus are unable to study the relationship between linguistic information and task performance. 

\section{Visualizing Relationship via Pareto Optimality \label{sec:3}}
We consider the relationship between linguistic information and task performance for two popular tasks in NLP, i.e., machine translation and language modeling.  
Let $\bm{x} = \{x_1, x_2, ..., x_N\}$ be a sentence and $\bm{s} = \{s_1, s_2, ..., s_N\}$ be the labels of the linguistic property of $\bm{x}$, where $s_i$ is the label for $x_i$, e.g., POS tag. On both tasks, a deep model typically encodes $\bm{x}$ into a hidden representation $\bm{h}$ with a sub-network $E$ parameterized by $\theta_e$: $ \bm{h} = E(\bm{x})$, and then uses another sub-network $D$ parameterized by $\theta_d$ to map $\bm{h}$ into an output.
\subsection{Background}

\paragraph{$\bm{h}$ and Loss in NMT}
An NMT architecture aims to output a target sentence $\bm{y} = \{y_1, y_2, ..., y_M\}$ for a given source sentence $\bm{x}$ according to $P(\bm{y}\mid \bm{x}; \theta)$~\cite{zaremba2014recurrent,vaswani2017attention}, where $\theta$ indicates a set of parameters of a sequence-to-sequence neural network, which contains an encoder $E$ and a decoder $D$. We define  $\bm{h}$ as the output of the encoder. 
To train $\theta$, the MLE loss is usually minimized on a training dataset.
For NMT, the loss is defined as following:
\begin{equation}
    L_{\theta}(\bm{x}, \bm{y}) = -\sum\limits_{j=1}^M\log P(y_j|\bm{x}, \bm{y}_{<j};\theta)\label{eq:mle}
\end{equation}
In our experiments, we consider two models, namely the LSTM~\cite{bahdanau2014neural} and Transformer~\cite{vaswani2017attention}. 

\paragraph{$\bm{h}$ and Loss in LM}
For language modeling task, a deep model typically generates a token $x_j$ based on $\bm{x}_{<j}$ according to $P(x_j|\bm{x}_{<j};\theta)$. Here the sub-networks $E$ is set as one hidden layer to encode $\bm{x}_{<j}$ into $\bm{h}_{<j}$ and $D$ is set as the sub-network to generate $x_j$ on top of $\bm{h}_{<j}$.
The parameter $\theta$ is optimized by the following MLE loss: $$L_{\theta}(\bm{x}) = -\sum\limits_{j=1}^N\log P(x_j| \bm{x}_{<j};\theta).\label{eq:mle-lm}$$ 

To make notations consistent for both NMT and LM, in the rest of this paper, we follow the form of Eq.~(\ref{eq:mle}) and re-write the $L_{\theta}(\bm{x})$ in LM as $L_{\theta}(\bm{x}, \bm{y})$, where $\bm{y}$ is a shifted version of $\bm{x}$, i.e., $\bm{y}=\{x_2, \cdots, x_N\}$. 


\paragraph{Encoded Information}
 Let $\mathrm{I}(\bm{h}, \bm{s})$ denote the linguistic information in the representation $\bm{h}$, \emph{i.e.}, the mutual information between $\bm{h}$ and the linguistic label $\bm{s}$. 
Since the probability $p(\bm{h}, \bm{s})$ is unknown, it is intractable to compute $\mathrm{I}(\bm{h}, \bm{s})$. Following \citet{pimentel2020information}, we approximately estimate $\mathrm{I}(\bm{h}, \bm{s})$ by using a probing model $q$ as follows:
\begin{equation}
    \begin{aligned}
\mathrm{I}(\bm{h}, \bm{s}) & = H(\bm{s}) - H(\bm{s}|\bm{h}) \\
&\approx \mathrm{H}(\bm{s}) - \min_{\theta_q} L_{\theta_q}(\bm{h}, \bm{s})\\
& = \mathrm{H}(\bm{s}) + \min_{\theta_q}\sum_{i}\log q(s_i|\bm{h};\theta_q)
    \end{aligned}
    \label{eq:info}
\end{equation}
where $H(\bm{s})$ is the entropy of linguistic labels, $H(\bm{s}|\bm{h})$ is the ideal cross entropy, and $L_{\theta_q}(\bm{h}, \bm{s})$ is the cross-entropy loss of the probe model $q$ parameterized by $\theta_q$.

\paragraph{Theory of Pareto Optimality}

Pareto optimality~\citep{mas1995microeconomic} is essentially involved in the multi-objective optimization problem. Suppose that we have $K$ different objectives $M_k$ to evaluate a parameter $\theta'$, \emph{i.e.}, 
\begin{equation}
    \arg\max_{\theta'}[M_1(\theta'); M_2(\theta'); ...; M_K(\theta')]\label{mt}.
\end{equation}
There are two important concepts in Pareto optimality as follows:\\
\textbf{Definition 1.} \textit{Pareto Optimal: A parameter $\theta^*$ is Pareto-optimal iff~ for any $\theta'$, the condition always holds that, $\forall i=1, ..., k,~M_i(\theta^*) \geq M_i(\theta')$ and $\exists j$ such that $M_j(\theta^*) > M_j(\theta')$.}\\
\textbf{Definition 2.} \textit{Pareto Frontier: The set of all Pareto-optimal parameters is called the Pareto frontier}.

\subsection{Viewpoint via Pareto Optimality}

\paragraph{Motivation}
Suppose $\theta$ is a given model parameter, $L(\theta)$ is its task performance on a test set, and $I(\theta)$ is the amount of linguistic information encoded in its hidden representation. Conventionally, if one can figure out a function $f$ such that $I=f(L)$ for any $\theta$, it is trivial to study their relationship by visualizing $f$. Unfortunately, for some complicated situations as illustrated in Figure~\ref{fig:illustration}, there does not exist such a function to represent the relationship between two variables due to a large number of many-to-many correspondences. 

\paragraph{Our Viewpoint} Pareto Optimality, a well-known criterion in economics~\citep{mas1995microeconomic}, is widely used to analyze the relationship among multiple variables in a complicated environment~\cite{chinchuluun2008Pareto}. 
In our context, it is also a powerful tool to reveal the relationship between the encoded linguistic information and task performance. Taking the Pareto Frontier in Figure \ref{fig:illustration} as an example, since the capacity of a model is fixed and linguistic information may compete with other kinds of information, capturing more linguistic information may reduce the amount of information from other sources that are also helpful for the model. Nevertheless, if increasing the amount of linguistic information constantly leads to performance gain, i.e., linguistic information is complimentary to translation, only one Pareto Optimal point would exist on the top right corner.

Therefore, in this paper, we propose to study the relationship between $I(\theta)$ and $L(\theta)$ from the viewpoint of Pareto Optimality. Our key idea is to take into account only Pareto-optimal models rather than all models like the conventional method. Thanks to the definition of Pareto optimality, there are no many-to-many correspondences between two variables along the Pareto frontier. Hence their relationship can be visualized by the trend of these frontier points, as shown in Figure~\ref{fig:illustration}. Taking Figure~\ref{fig:illustration} as an example, to answer the questions mentioned before,  we can see that adding more information is possible to increase the task performance comparing with a standard model. According to this viewpoint, the core challenge is how to obtain a set of models which are Pareto optimality on a test dataset.

It is natural to employ a \textit{heuristic} method to approximately obtain the Pareto-optimal models as following. We can first randomly select a number of checkpoints during the standard training and probe each checkpoint by optimizing its corresponding probing model $q$, as shown in Eq~\eqref{eq:info}. Second, we can record the task performances and the amounts of linguistic information of the selected models on a test set. Finally, we can find the Pareto-optimal points and obtain the Pareto frontier. However, when using this method in our experiments, we find the amounts of encoded linguistic information for all checkpoints are similar and the the task performances of those checkpoints are worse than the optimal model. Hence, in the next section, a new method will be presented to approximately derive the Pareto-optimal models. 


\section{Methodology}

\subsection{Multi-Objective Optimization}
To study the relationship between linguistic information and task performance, our goal is to obtain a set of models $\theta$ which are Pareto optimal on test data in terms of both objectives. Inspired by statistical learning theory~\cite{vapnik1999overview}, we propose an approach by optimizing the Pareto-optimal models towards both objectives on a given training dataset, which are expected to generalize well on unseen test data, i.e., these models are Pareto optimal on unseen test data. 
Formally, Our approach can be formulated as the following multi-objective optimization problem:
    \begin{equation}
       \arg\min_\theta [L_{\theta}(\bm{x}, \bm{y}); -\mathrm{I}(\bm{h}, \bm{s})]\label{eq:add-org} \\
    \end{equation}
\noindent where minimizing $L_{\theta}(\bm{x}, \bm{y})$ aims to promote the task performance and maximizing $\mathrm{I}(\bm{h}, \bm{s})$ encourages a model to encode more linguistic information in the representation. Once we obtain a set of Pareto-optimal models, we can observe how increasing the encoded linguistic information affects the variance of task performance. 

To further study how reducing the encoded linguistic information affects task performance, we optimize a similar multi-objective problem:
    \begin{equation}
       \arg\min_\theta [L_{\theta}(\bm{x}, \bm{y}); \mathrm{I}(\bm{h}, \bm{s})]\label{eq:rmv-org}
    \end{equation}
The only difference between Eq.~\eqref{eq:add-org} and Eq.~\eqref{eq:rmv-org} is that the former maximizes $\mathrm{I}(\bm{h}, \bm{s})$ while the latter minimizes $\mathrm{I}(\bm{h}, \bm{s})$.

Since $H(\bm{s})$ is a constant term, we can plug Eq.~\eqref{eq:info} into the above two equations and obtain the following reduced multi-objective problems:
    \begin{align}
       & \arg\min_\theta [L_{\theta}(\bm{x}, \bm{y}); \min_{\theta_q} L_{\theta_q}(\bm{h}, \bm{s})]\label{eq:add} \\
       & \arg\min_\theta [L_{\theta}(\bm{x}, \bm{y}); -\min_{\theta_q} L_{\theta_q}(\bm{h}, \bm{s})]\label{eq:rmv}
    \end{align}
Notice that in the above equations, $\min\limits_{\theta_q} L_{\theta_q}(\bm{h}, \bm{s})$ resembles a conventional probing if $\bm{h}$ is a fixed representation. However, unlike the standard probing applied on top of a fixed $\bm{h}$ determined by the standard model, here $\bm{h}$ is the representation obtained from a encoder $E$ parameterized by $\theta_e$. It is also worth noting that the Pareto frontiers obtained from the Eq. (\ref{eq:add}) and (\ref{eq:rmv}) are independent, although they have a similar measurement, because the Pareto Optimal is only valid for the same objective.

\subsection{Optimization Algorithm}
To solve the above multi-objective problems, we leverage the linear-combination method to find a set of solutions, and then filter the non-Pareto-optimal points from the set to get the Pareto frontier. The details of our algorithm are shown below.

\paragraph{Optimization Process}
Since the detailed optimization method for Eq.~\eqref{eq:add} is similar to that for Eq.~\eqref{eq:rmv}, in the following we take Eq.~\eqref{eq:add} as an example to describe the optimization method. Inspired by \citep{duh2012learning}, we employ a two-step strategy for optimization to find the Pareto frontier to address the multi-objective problems. 
\begin{figure}
\begin{center}
   \includegraphics[width=0.85\linewidth]{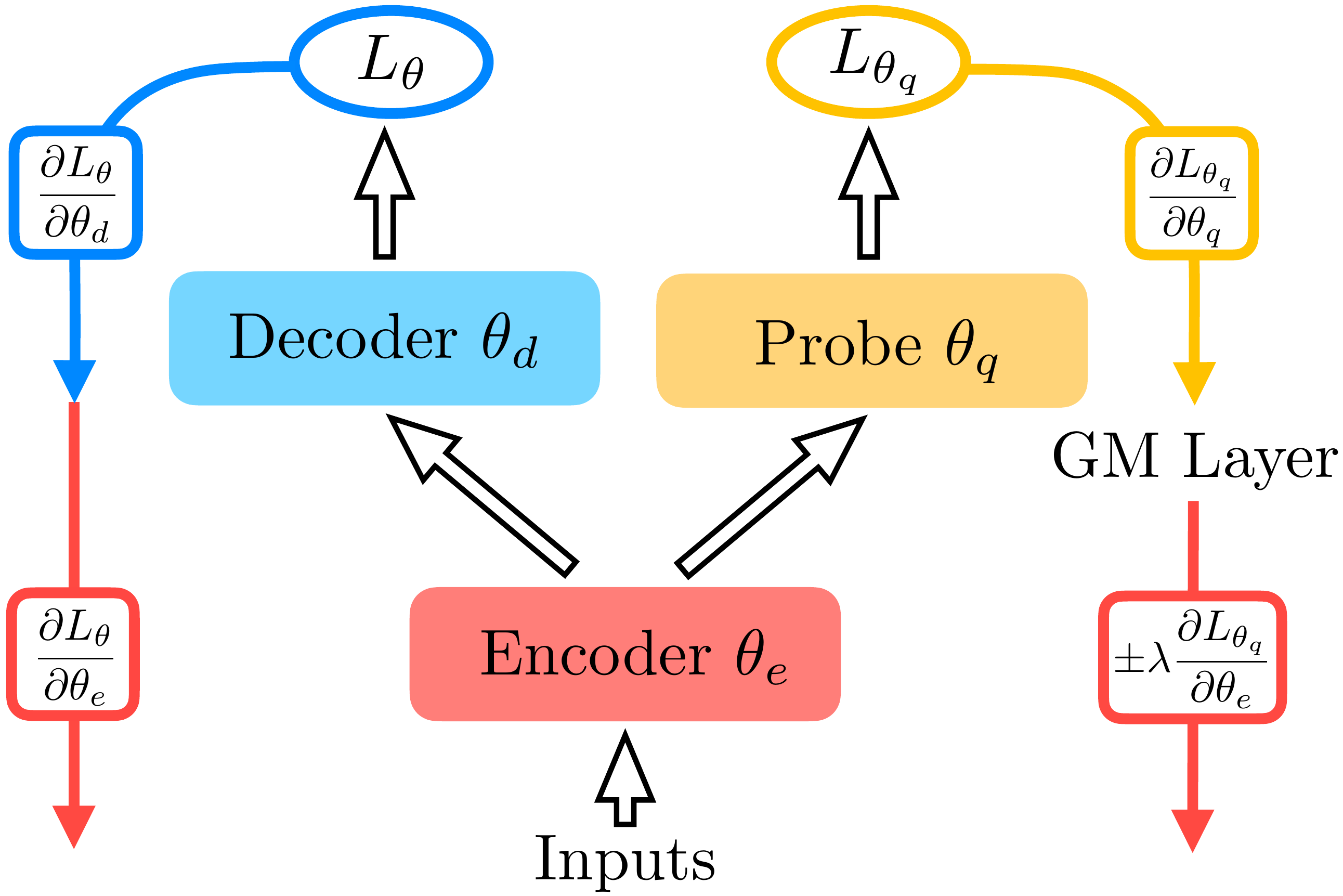}
\end{center}
\vspace{-2ex}
   \caption{Overview of our multi-objective optimization method, where $L_y = L_{\theta}(\bm{x}, \bm{y})$ and $L_{\theta_q} = L_{\theta_q}(\bm{h}, \bm{s})$. In the back propagation, the GM Layer multiplies the gradient by $\pm \lambda$, i.e., $\lambda$ for Eq.~\eqref{eq:add} and $-\lambda$ for Eq.~\eqref{eq:rmv}.}
\label{model}    \vspace{-2ex}
\end{figure}

In the first step, we adopt an method to find the Pareto-optimal solutions to the problem. There are several different methods to solve the problem, such as linear-combination, PMO~\citep{duh2012learning} and APStar~\citep{martinez2020minimax}. In this work, we adopt the linear-combination method because of its simplicity. Specifically, we select a coefficient set $\{\lambda_k \mid \lambda_k > 0\}$ and minimize the following interpolating function for each coefficient $\lambda_k$:~\footnote{Eq.~\eqref{eq:interpolating} is similar to the loss of standard multiple task learning (MTL)~\cite{dong2015multi,lee-etal-2020-discrepancy}, $\arg\min_{\theta, \theta_q} \big(L_{\theta}(\bm{x}, \bm{y}) \pm \lambda_k L_{\theta_q}(\bm{h},\bm{s})\big)$. 
However, the solutions to the loss are weaker than our solutions according to Pareto optimality, and it can not remove linguistic information in our preliminary experiments.}
\begin{equation}
    \arg\min_{\theta} \big(L_{\theta}(\bm{x}, \bm{y}) + \lambda_k \min_{\theta_q}L_{\theta_q}(\bm{h},\bm{s})\big)
    \label{eq:interpolating}
\end{equation}
Notice that the first term of the loss function $L_{\theta}(\bm{x}, \bm{y})$ is the function of both encoder parameters $\theta_e$ and decoder parameters $\theta_d$, while the second term $\min_{\theta_q} L_{\theta_q}(\bm{h}, \bm{s})$ is only the function of $\theta_e$. 
Therefore, when minimizing Eq.\eqref{eq:interpolating}, we apply a Gradient-Multiple (GM) Layer on the representations before inputting it into the probe model. As shown in Fig.~\ref{model}, in the forward propagation, the GM Layer acts as an identity transform, while in the backward propagation, the GM Layer multiples the gradient by $\pm\lambda$ and passes it to the preceding layers. Note that when the multiplier is $-\lambda$, the GM Layer is the same as Gradient Reversal Layer \citep{ganin2015unsupervised}.

Suppose $\{\theta^*_k\mid \theta^*_k > 0 \}$ is the minimized solution set for Eq.~\eqref{eq:interpolating}. 
In the second step, to get more accurate solutions, we filter the non-Pareto-optimal points of the solution set obtained by $\{\theta^*_k\mid \theta^*_k > 0 \}$. Finally, we get the Pareto frontier to the multi-objective problem according to the definition of Pareto optimality.

\begin{algorithm}[t]
	\caption{Optimization Algorithm}
	\begin{algorithmic}[1]
       \INPUT
       $\Lambda = \{\lambda_k\} $, learning rate $\eta$
        \OUTPUT Pareto frontier set $ \mathcal{P} = \{ \langle \theta_e^i, \theta_d^i \theta_q^i \rangle \}$
    \State $\mathcal{M} = \{\}$ \Comment{empty model set}
    \For{$\lambda_k \in \Lambda $}\Comment{minimize Eq.~ (\ref{eq:interpolating})}
      \State Random initialize $\theta_e^k, \theta_d^k$, and $\theta_q^k$  
      \While {convergence}
        \State $\theta_e^k = \theta_e^k - \eta (\frac{\partial L_\theta(x, y)}{\partial \theta_e} + \lambda_k\frac{\partial L_{\theta_q}(\bm{s}, \bm{h})}{\partial \theta_e})$ \Comment{$+\lambda_k$ is for Eq.~\eqref{eq:add} and changing it to $-\lambda_k$ would optimize Eq.~\eqref{eq:rmv}}
        \State $\theta_d^k = \theta_d^k - \eta \frac{\partial L_\theta(x, y)}{\partial \theta_d}$
        \State $\theta_q^k = \theta_q^k - \eta \frac{\partial L_{\theta_q}(\bm{s}, \bm{h})}{\partial \theta_q}$
      \EndWhile
      \State Re-train a probe model $\theta_{q\prime}^k$ based on fixed encoder $\theta_e$
      \State Add $\langle \theta_e^k, \theta_d^k, \theta_{q\prime}^k \rangle$ into $\mathcal{M}$
    \EndFor
    \State $\mathcal{P} = \{\}$ \Comment{Pareto frontier set}
    \ForAll{$\langle \theta_e^k, \theta_d^k, \theta_{q\prime}^k \rangle \in \mathcal{M}$}
      \If{\texttt{IsParetoOptimal}$ (\theta_e^k, \theta_d^k, \theta_{q\prime}^k )$}
        \State Add $\langle \theta_e^k, \theta_d^k, \theta_{q\prime}^k \rangle $ into $\mathcal{P}$
      \EndIf
    \EndFor
	\end{algorithmic}
\label{alg}
\end{algorithm}

\paragraph{Detailed Algorithm}
The overall optimization algorithm regarding to Eq.~\eqref{eq:add} is shown in Algorithm 1.   
Theoretically, when minimizing Eq.~\eqref{eq:interpolating}, in every step updating $\theta$, we should retrain the probe model $\theta_q$ to minimize $L_{\theta_q}(\bm{h},\bm{s})$ in for many steps, in order to estimate $\mathrm{H}(\bm{s}|\bm{h})$ precisely. However, this is time-consuming and inefficient. Instead, after updating $\theta$, we update $\theta_q$ only by one step (see line 7 Algorithm 1). Empirically, we find that optimization in this way has been very effective.

In addition, as is mentioned by \citet{elazar2018adversarial}, information leakage may occur when minimizing the mutual information. Therefore, after the training process is finished, we fix the deep model and retrain another probe model to estimate $\mathrm{H}(\bm{s}|\bm{h})$ more precisely (line 9 in Algorithm 1). When maximizing the mutual information, we find there is no difference between $\mathrm{H}(\bm{s}|\bm{h})$ estimated by jointly trained or retrained probe models.

\section{Experimental Settings}

\begin{figure*}[htp]
\begin{center}
   \includegraphics[width=1\linewidth]{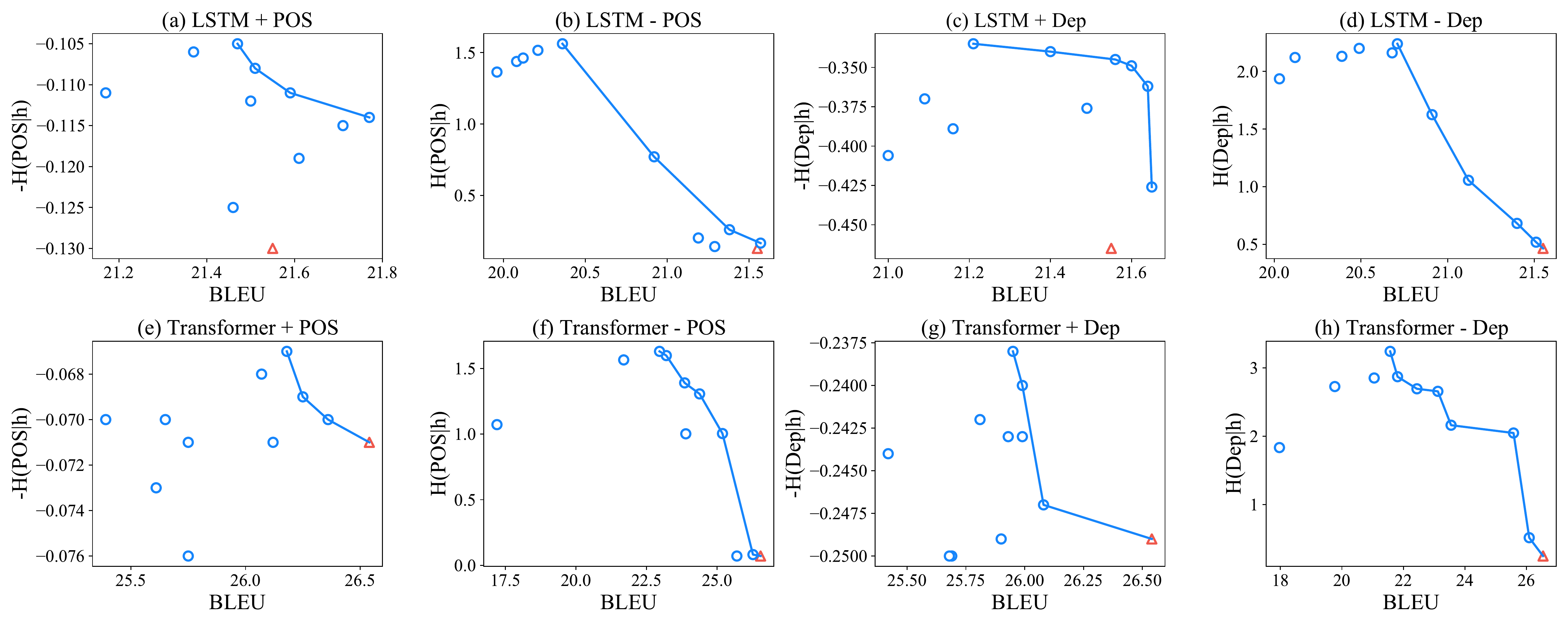}
\end{center}    \vspace{-2ex}
   \caption{Experiments on WMT14 corpus. Triangle ($\bigtriangleup$) denotes the model trained by minimizing MLE loss, circle ($\bigcirc$) denotes the models obtained by our method, and the models on the line (---) denotes the Pareto frontier.
   }     \vspace{-2ex}
\label{wmt14}
\end{figure*}

\subsection{Dataset} We conduct experiments on both machine translation and language modeling tasks. For machine translation, we conduct the experiments on En $\Rightarrow$ De and Zh $\Rightarrow$ En translation tasks. For En $\Rightarrow$ De task, we use WMT14 corpus which contains 4M sentence pairs. For Zh $\Rightarrow$ En task, we use LDC corpus which consists of 1.25M sentence pairs, and we choose NIST02 as our validation set, and NIST06 as our test set. For language modeling task, we use Penn Treebank\footnote{\url{https://deepai.org/dataset/penn-treebank}} dataset. We preprocess our data using byte-pair encoding \citep{sennrich2015neural} and keep all tokens in the vocabulary. For machine translation, we use case-insensitive 4-gram BLEU score \citep{papineni2002bleu} to measure the task performance, which is proved to be positively correlated well with the MLE loss \cite{lee2020discrepancy}; For language modeling, we directly use the 
MLE loss to evaluate the task performance.

\subsection{Linguistic Properties} For machine translation, we study part-of-speech (POS) and dependency labels in this work. Since there are no gold labels for the MT datasets, we use Stanza toolkit\footnote{\url{https://github.com/stanfordnlp/stanza}} \citep{qi2020stanza} to annotate source sentences and use the pseudo labels for running our algorithm, following~\citet{sennrich2016linguistic,li2018target}.
We clean the labels and remove the sentences that fail to be parsed by Stanza from the dataset. To study whether all kinds of linguistic information are critical for neural models, we also investigate the phonetic information on the language modeling task. More precisely, the probing model needs to predict the first character of the International Phonetic Alphabet of each word.\footnote{For example, given the input sentence "This dog is so cute", the probing model is asked to predict "\textipa{ð} \textipa{d} \textipa{I} \textipa{s} \textipa{k}".} We get the labels with the open source toolkit English-to-IPA\footnote{\url{https://github.com/mphilli/English-to-IPA}}. We use mutual information $\mathrm{I}(\bm{h}, \bm{s}) = \mathrm{H}(\bm{s}) - \mathrm{H}(\bm{s}|\bm{h})$ to evaluate the amount of information in the representations. Since $\mathrm{H}(\bm{s})$ is a constant, we only compare $\mathrm{H}(\bm{s}|\bm{h})$ in the experiments. Note that $\mathrm{H}(\bm{s}|\bm{h})$ is estimated by our probe model $q$.

\subsection{Implementation Details} All of our models are implemented with Fairseq\footnote{\url{https://github.com/pytorch/fairseq}} \citep{ott2019fairseq}. For NMT experiments, our LSTM model consists of a bi-directional 2-layer encoder with 256 hidden units, and a 2-layer decoder with 512 hidden units, and the probe model is a 2-layer MLP with 512 hidden units. Our Transformer model consists of a 6-layer encoder and a 6-layer decoder, whose hyper-parameters are the same as the base model in \citep{vaswani2017attention}, and the probe model is a 6-layer transformer encoder. For LM experiments, our model is a 2-layer LSTM with 256 hidden states, and the probe model is a 2-layer MLP with 256 hidden states. More training details about our models are shown in appendix \ref{appendix:train}.


\begin{figure}[]
    \begin{center}
        \includegraphics[width=0.8\linewidth]{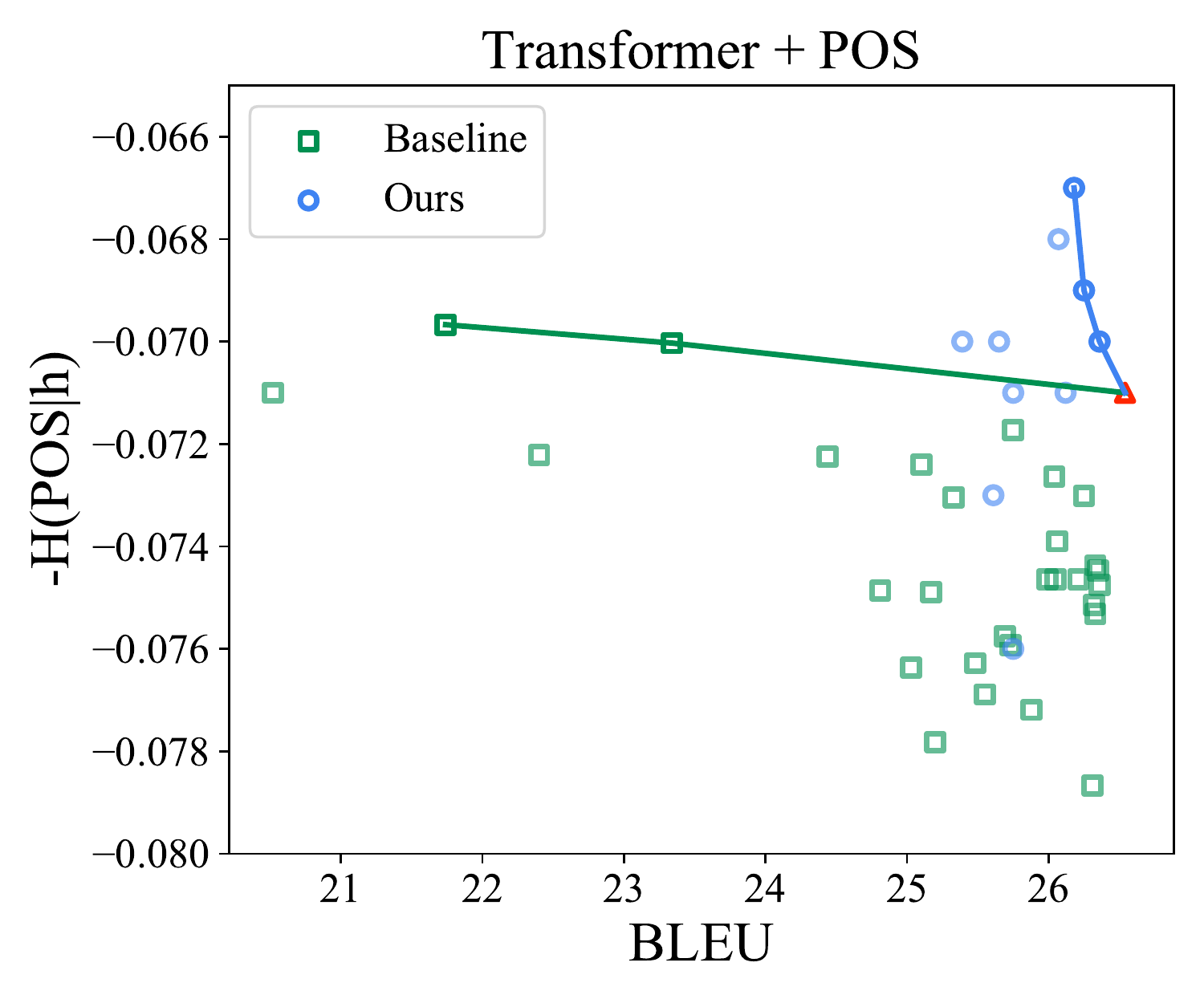}
    \end{center}
    \vspace{-2ex}
        \caption{Comparison with baseline method. Triangle ($\bigtriangleup$) denotes the standard model by minimizing MLE loss. The \textcolor{green}{green line} and \textcolor{blue}{blue line} are frontiers got from baseline method and our method respectively. \label{fig:comparison}}    \vspace{-2ex}
\end{figure}

\begin{figure*}[htp]
\begin{center}
   \includegraphics[width=1\linewidth]{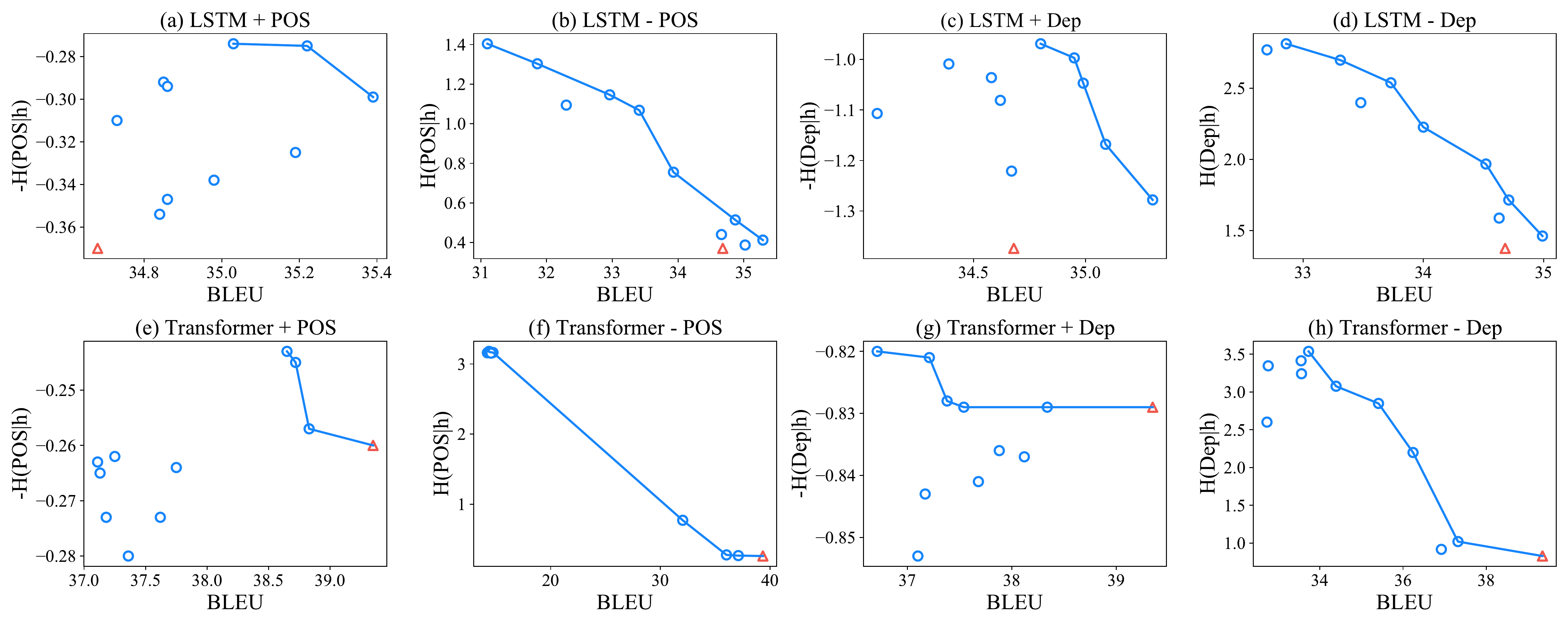}
\end{center}
\vspace{-2ex}
   \caption{Experimental results on LDC corpus. The format is the same as Fig.~\ref{wmt14}}
\label{ldc}    \vspace{-2ex}
\end{figure*}

\section{Experiment Results}
In the following experiments, "{\bf Model + Property}", e.g., "Transformer+Pos", which is corresponding to Eq.~\ref{eq:add-org} and studies how adding the linguistic properties information affects the task performance. Instead, "{\bf Model - Property}", e.g., "Transformer-Pos", which is corresponding to Eq.~\ref{eq:rmv-org} and studies how removing the linguistic properties information affects the task performance. It is worth noting that merging the two frontiers of \textbf{+ Property} and \textbf{- Property} together would lead to trivial results, because Pareto Optimal points of the \textbf{+ Property} are more likely to dominate. However, we think the frontier of \textbf{- Property} is helpful for answering the question that whether reducing the encoded linguistic information would affect the model performance. Therefore, we plot the Pareto frontiers for the two objectives independently.
\subsection{Soundness of Methodology}

The heuristic method mentioned before can be considered as a simple and straightforward baseline method to measure the relationship. To set up this baseline, we firstly save some checkpoints every 1,000 steps when training a standard model. Second, we randomly sample 30 checkpoints for probing and plot a scatter diagram in terms of BLEU and encoded linguistic information.

As shown in Figure \ref{fig:comparison}, we compare our proposed method with the heuristic method in the setting of "Transformer+Pos". Comparing with the baseline method, the frontier obtained from our method is better: for each model explored by baseline, there exists at least one model explored by our method whose two objectives, i.e., encoded linguistic information and BLEU score, are larger. The main reason is that the objective of baseline method only considers the task performance and most checkpoints contain similar encoded linguistic information. Therefore, the models optimized by our multi-objective method is more close to the globally Pareto-optimal points~\footnote{It is worth mentioning that there are no algorithms to guarantee globally Pareto-optimal solutions in our scenario on the training data. Although the globally Pareto-optimal solutions are unknown, our solutions are definitely more close to them than the solutions by baseline.}, making the revealed relationship between encoded linguistic information and task performance more reliable. Therefore, in the next subsection, our proposed method will be used to visualize the relationship between encoded linguistic information and task performance for neural models.

\subsection{Visualization Results}

\paragraph{Results on NMT} 
The results of machine translation on the WMT dataset are shown in Figure~\ref{wmt14}. For LSTM based NMT, we observe that the standard model, i.e., the $\bigtriangleup$ in Figure~\ref{wmt14},  is not in the Pareto frontier in Figure~\ref{wmt14} (a,c). In other words, when adding linguistic information into the LSTM model, it is possible to obtain a model which contains more POS or DEP information and meanwhile leads to better BLEU score than the standard model by standard training. In contrast, for Transformer based NMT, the standard model is in the Pareto frontier, as shown in Figure~\ref{wmt14} (e,g). This finding provides an explanation to the fact in NMT: many efforts~\cite{DBLP:journals/corr/LuongLSVK15,DBLP:conf/wmt/NadejdeRSDJKB17,DBLP:conf/emnlp/BastingsTAMS17,DBLP:conf/emnlp/HashimotoT17,DBLP:conf/acl/EriguchiTC17} have been devoted to improve the LSTM based  NMT architecture by explicitly modeling linguistic properties,  
but few have been done on Transformer based NMT \cite{mcdonald2021syntax, currey2019incorporating}. In addition, when removing the linguistic information from LSTM or Transformer, the standard model is very close to the lower right of Pareto frontier, or even at the frontier, as shown in Figure~\ref{wmt14} (b,d,f,h). This result shows that removing linguistic information always hurts the performance of NMT models for both LSTM and Transformer, indicating that encoding POS and DEP information is important for NMT task. Similar trends are observed on the LDC datasets, as shown in Figure~\ref{ldc}. More details about the effect of randomness on our approach are shown in appendix \ref{sec:rand}.

\begin{figure}
\begin{center}
   \includegraphics[width=1\linewidth]{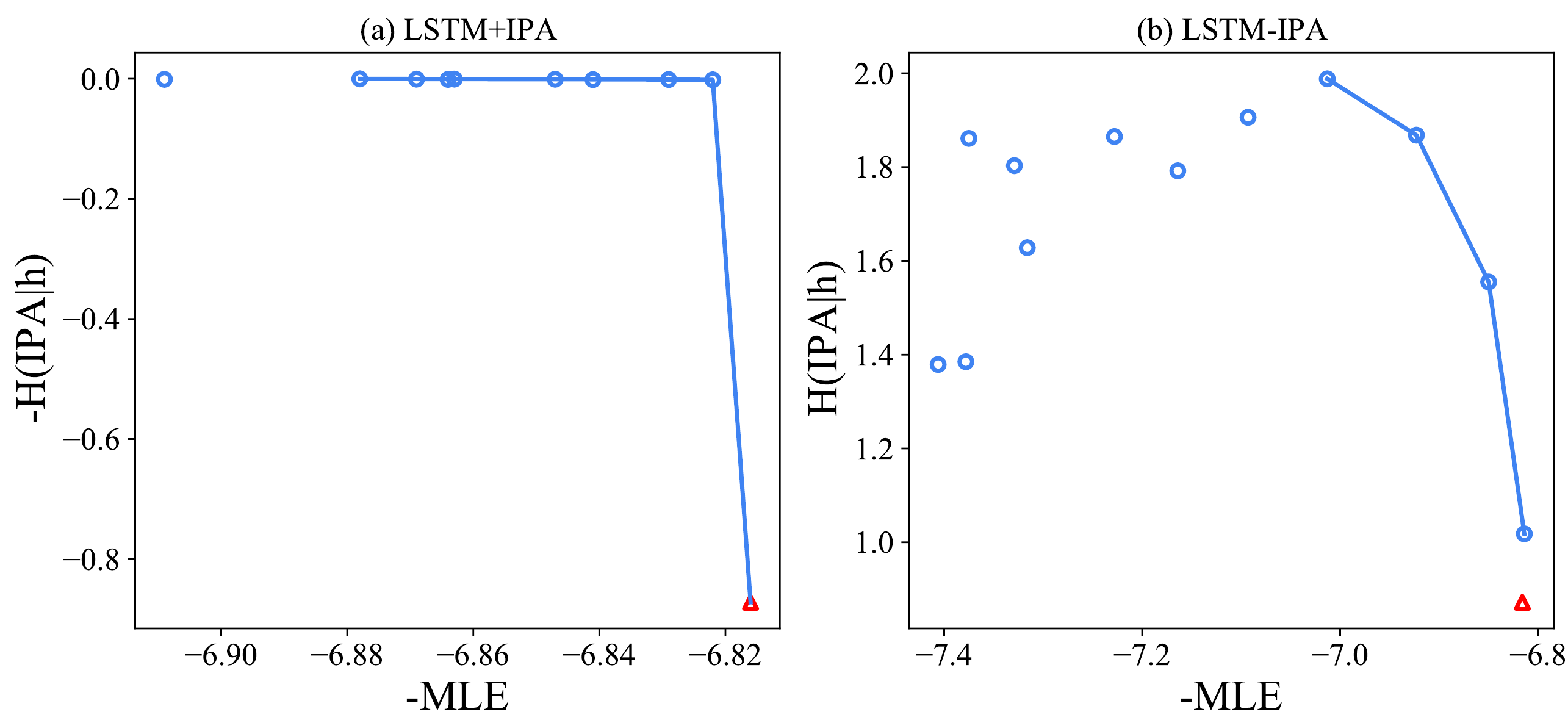}
\end{center}
\vspace{-2ex}
   \caption{Experimental results on the PTB dataset. 
   }    \vspace{-2ex}
  \label{fig:ipa_lstm}
\end{figure}

\paragraph{Results on LM} 
Above experiments have shown that both syntactic information are important for NMT models, and then a natural question is whether all kinds of linguistic information are important for neural models.   
To answer this question, we propose to investigate the influence of phonetic information on a language model. Figure~\ref{fig:ipa_lstm} depicts the relationship between encoded phonetic information and task performance for an LSTM based language model. In Figure~\ref{fig:ipa_lstm} (a), we find that the performances of Pareto-optimal models drop slightly when forcing an LSTM model to encode more phonetic information. Besides, as the Pareto-frontier shown in Figure~\ref{fig:ipa_lstm} (b), removing phonetic information from an LSTM model only leads to a slight change in performance. These experiments demonstrate that the encoded phonetic information may be not that critical for an LSTM based language model. This finding suggest that not all kinds of linguistic information are crucial for LSTM based LM and it is not promising to further improve language modeling with phonetic information. 

\section{Conclusion}

This paper aims to study the relationship between linguistic information and the task performance and proposes a new viewpoint inspired by the criterion of Pareto Optimality. We formulate this goal as a multi-objective problem and present an effective method to address the problem by leveraging the theory of Pareto optimality. 
We conduct experiments on both MT and LM tasks and study their performance with respect to linguistic information sources. Experimental results show that the presented approach is more plausible than a baseline method in the sense that it explores better models in terms of both encoded linguistic information and task performance. In addition, we obtain some valuable findings as follows: i) syntactic information encoded by NMT models is important for MT task and reducing it leads to sharply decreased performance; ii) the standard NMT model obtained by minimizing MLE loss is Pareto-optimal for Transformer but it is not the case for LSTM based NMT; iii) reducing the phonetic information encoded by LM models only leads to slight performance drop. 

\section*{Acknowledgement}
We would like to thank the anonymous reviewers for their constructive comments. 
L. Liu is the corresponding author. 

\bibliography{anthology,custom}

\begin{thebibliography}{53}
\expandafter\ifx\csname natexlab\endcsname\relax\def\natexlab#1{#1}\fi

\bibitem[{Adi et~al.(2017)Adi, Kermany, Belinkov, Lavi, and
  Goldberg}]{adi2016fine}
Yossi Adi, Einat Kermany, Yonatan Belinkov, Ofer Lavi, and Yoav Goldberg. 2017.
\newblock Fine-grained analysis of sentence embeddings using auxiliary
  prediction tasks.
\newblock In \emph{5th International Conference on Learning Representations,
  {ICLR} 2017, Toulon, France, April 24-26, 2017, Conference Track
  Proceedings}. OpenReview.net.

\bibitem[{Alt et~al.(2020)Alt, Gabryszak, and Hennig}]{alt2020probing}
Christoph Alt, Aleksandra Gabryszak, and Leonhard Hennig. 2020.
\newblock \href {https://doi.org/10.18653/v1/2020.acl-main.140} {Probing
  linguistic features of sentence-level representations in neural relation
  extraction}.
\newblock In \emph{Proceedings of the 58th Annual Meeting of the Association
  for Computational Linguistics}, pages 1534--1545, Online. Association for
  Computational Linguistics.

\bibitem[{Bahdanau et~al.(2015)Bahdanau, Cho, and Bengio}]{bahdanau2014neural}
Dzmitry Bahdanau, Kyunghyun Cho, and Yoshua Bengio. 2015.
\newblock Neural machine translation by jointly learning to align and
  translate.
\newblock In \emph{3rd International Conference on Learning Representations,
  {ICLR} 2015, San Diego, CA, USA, May 7-9, 2015, Conference Track
  Proceedings}.

\bibitem[{Bastings et~al.(2017)Bastings, Titov, Aziz, Marcheggiani, and
  Sima{'}an}]{DBLP:conf/emnlp/BastingsTAMS17}
Jasmijn Bastings, Ivan Titov, Wilker Aziz, Diego Marcheggiani, and Khalil
  Sima{'}an. 2017.
\newblock \href {https://doi.org/10.18653/v1/D17-1209} {Graph convolutional
  encoders for syntax-aware neural machine translation}.
\newblock In \emph{Proceedings of the 2017 Conference on Empirical Methods in
  Natural Language Processing}, pages 1957--1967, Copenhagen, Denmark.
  Association for Computational Linguistics.

\bibitem[{Beckman et~al.(2002)Beckman, Formby, Smith, and
  Zheng}]{beckman2002envy}
Steven~R Beckman, John~P Formby, W~James Smith, and Buhong Zheng. 2002.
\newblock Envy, malice and pareto efficiency: An experimental examination.
\newblock \emph{Social Choice and Welfare}, 19(2):349--367.

\bibitem[{Cao et~al.(2021)Cao, Sanh, and Rush}]{cao-etal-2021-low}
Steven Cao, Victor Sanh, and Alexander Rush. 2021.
\newblock \href {https://doi.org/10.18653/v1/2021.naacl-main.74}
  {Low-complexity probing via finding subnetworks}.
\newblock In \emph{Proceedings of the 2021 Conference of the North American
  Chapter of the Association for Computational Linguistics: Human Language
  Technologies}, pages 960--966, Online. Association for Computational
  Linguistics.

\bibitem[{Chen et~al.(2018)Chen, Firat, Bapna, Johnson, Macherey, Foster,
  Jones, Schuster, Shazeer, Parmar, Vaswani, Uszkoreit, Kaiser, Chen, Wu, and
  Hughes}]{chen-etal-2018-best}
Mia~Xu Chen, Orhan Firat, Ankur Bapna, Melvin Johnson, Wolfgang Macherey,
  George Foster, Llion Jones, Mike Schuster, Noam Shazeer, Niki Parmar, Ashish
  Vaswani, Jakob Uszkoreit, Lukasz Kaiser, Zhifeng Chen, Yonghui Wu, and
  Macduff Hughes. 2018.
\newblock \href {https://doi.org/10.18653/v1/P18-1008} {The best of both
  worlds: Combining recent advances in neural machine translation}.
\newblock In \emph{Proceedings of the 56th Annual Meeting of the Association
  for Computational Linguistics (Volume 1: Long Papers)}, pages 76--86,
  Melbourne, Australia. Association for Computational Linguistics.

\bibitem[{Chinchuluun et~al.(2008)Chinchuluun, Pardalos, Migdalas, and
  Pitsoulis}]{chinchuluun2008Pareto}
Altannar Chinchuluun, Panos~M Pardalos, Athanasios Migdalas, and Leonidas
  Pitsoulis. 2008.
\newblock \emph{Pareto optimality, game theory and equilibria}.
\newblock Springer.

\bibitem[{Conneau et~al.(2018)Conneau, Kruszewski, Lample, Barrault, and
  Baroni}]{conneau2018you}
Alexis Conneau, German Kruszewski, Guillaume Lample, Lo{\"\i}c Barrault, and
  Marco Baroni. 2018.
\newblock \href {https://doi.org/10.18653/v1/P18-1198} {What you can cram into
  a single {\$}{\&}!{\#}* vector: Probing sentence embeddings for linguistic
  properties}.
\newblock In \emph{Proceedings of the 56th Annual Meeting of the Association
  for Computational Linguistics}, pages 2126--2136, Melbourne, Australia.
  Association for Computational Linguistics.

\bibitem[{Currey and Heafield(2019)}]{currey2019incorporating}
Anna Currey and Kenneth Heafield. 2019.
\newblock \href {https://doi.org/10.18653/v1/W19-5203} {Incorporating source
  syntax into transformer-based neural machine translation}.
\newblock In \emph{Proceedings of the Fourth Conference on Machine
  Translation}, pages 24--33, Florence, Italy. Association for Computational
  Linguistics.

\bibitem[{Devlin et~al.(2019)Devlin, Chang, Lee, and
  Toutanova}]{devlin2018bert}
Jacob Devlin, Ming-Wei Chang, Kenton Lee, and Kristina Toutanova. 2019.
\newblock \href {https://doi.org/10.18653/v1/N19-1423} {{BERT}: Pre-training of
  deep bidirectional transformers for language understanding}.
\newblock In \emph{Proceedings of the 2019 Conference of the North {A}merican
  Chapter of the Association for Computational Linguistics: Human Language
  Technologies}, pages 4171--4186, Minneapolis, Minnesota. Association for
  Computational Linguistics.

\bibitem[{Ding et~al.(2017)Ding, Liu, Luan, and
  Sun}]{ding-etal-2017-visualizing}
Yanzhuo Ding, Yang Liu, Huanbo Luan, and Maosong Sun. 2017.
\newblock \href {https://doi.org/10.18653/v1/P17-1106} {Visualizing and
  understanding neural machine translation}.
\newblock In \emph{Proceedings of the 55th Annual Meeting of the Association
  for Computational Linguistics (Volume 1: Long Papers)}, pages 1150--1159,
  Vancouver, Canada. Association for Computational Linguistics.

\bibitem[{Dong et~al.(2015)Dong, Wu, He, Yu, and Wang}]{dong2015multi}
Daxiang Dong, Hua Wu, Wei He, Dianhai Yu, and Haifeng Wang. 2015.
\newblock \href {https://doi.org/10.3115/v1/P15-1166} {Multi-task learning for
  multiple language translation}.
\newblock In \emph{Proceedings of the 53rd Annual Meeting of the Association
  for Computational Linguistics and the 7th International Joint Conference on
  Natural Language Processing (Volume 1: Long Papers)}, pages 1723--1732,
  Beijing, China. Association for Computational Linguistics.

\bibitem[{Duh et~al.(2012)Duh, Sudoh, Wu, Tsukada, and
  Nagata}]{duh2012learning}
Kevin Duh, Katsuhito Sudoh, Xianchao Wu, Hajime Tsukada, and Masaaki Nagata.
  2012.
\newblock \href {https://aclanthology.org/P12-1001} {Learning to translate with
  multiple objectives}.
\newblock In \emph{Proceedings of the 50th Annual Meeting of the Association
  for Computational Linguistics (Volume 1: Long Papers)}, pages 1--10, Jeju
  Island, Korea. Association for Computational Linguistics.

\bibitem[{Elazar and Goldberg(2018)}]{elazar2018adversarial}
Yanai Elazar and Yoav Goldberg. 2018.
\newblock \href {https://doi.org/10.18653/v1/D18-1002} {Adversarial removal of
  demographic attributes from text data}.
\newblock In \emph{Proceedings of the 2018 Conference on Empirical Methods in
  Natural Language Processing}, pages 11--21, Brussels, Belgium. Association
  for Computational Linguistics.

\bibitem[{Elazar et~al.(2020)Elazar, Ravfogel, Jacovi, and
  Goldberg}]{elazar2020bert}
Yanai Elazar, Shauli Ravfogel, Alon Jacovi, and Yoav Goldberg. 2020.
\newblock When bert forgets how to pos: Amnesic probing of linguistic
  properties and mlm predictions.
\newblock \emph{arXiv preprint arXiv:2006.00995}.

\bibitem[{Eriguchi et~al.(2017)Eriguchi, Tsuruoka, and
  Cho}]{DBLP:conf/acl/EriguchiTC17}
Akiko Eriguchi, Yoshimasa Tsuruoka, and Kyunghyun Cho. 2017.
\newblock \href {https://doi.org/10.18653/v1/P17-2012} {Learning to parse and
  translate improves neural machine translation}.
\newblock In \emph{Proceedings of the 55th Annual Meeting of the Association
  for Computational Linguistics}, pages 72--78, Vancouver, Canada. Association
  for Computational Linguistics.

\bibitem[{Ganin and Lempitsky(2015)}]{ganin2015unsupervised}
Yaroslav Ganin and Victor~S. Lempitsky. 2015.
\newblock Unsupervised domain adaptation by backpropagation.
\newblock In \emph{Proceedings of the 32nd International Conference on Machine
  Learning, {ICML} 2015, Lille, France, 6-11 July 2015}, volume~37 of
  \emph{{JMLR} Workshop and Conference Proceedings}, pages 1180--1189.
  JMLR.org.

\bibitem[{Goodfellow et~al.(2014)Goodfellow, Pouget{-}Abadie, Mirza, Xu,
  Warde{-}Farley, Ozair, Courville, and Bengio}]{goodfellow2014generative}
Ian~J. Goodfellow, Jean Pouget{-}Abadie, Mehdi Mirza, Bing Xu, David
  Warde{-}Farley, Sherjil Ozair, Aaron~C. Courville, and Yoshua Bengio. 2014.
\newblock Generative adversarial nets.
\newblock In \emph{Advances in Neural Information Processing Systems 27: Annual
  Conference on Neural Information Processing Systems 2014, December 8-13 2014,
  Montreal, Quebec, Canada}, pages 2672--2680.

\bibitem[{Greenwald and Stiglitz(1986)}]{greenwald1986externalities}
Bruce~C Greenwald and Joseph~E Stiglitz. 1986.
\newblock Externalities in economies with imperfect information and incomplete
  markets.
\newblock \emph{The quarterly journal of economics}, 101(2):229--264.

\bibitem[{Hashimoto and Tsuruoka(2017)}]{DBLP:conf/emnlp/HashimotoT17}
Kazuma Hashimoto and Yoshimasa Tsuruoka. 2017.
\newblock \href {https://doi.org/10.18653/v1/D17-1012} {Neural machine
  translation with source-side latent graph parsing}.
\newblock In \emph{Proceedings of the 2017 Conference on Empirical Methods in
  Natural Language Processing}, pages 125--135, Copenhagen, Denmark.
  Association for Computational Linguistics.

\bibitem[{Hewitt and Liang(2019)}]{hewitt2019designing}
John Hewitt and Percy Liang. 2019.
\newblock \href {https://doi.org/10.18653/v1/D19-1275} {Designing and
  interpreting probes with control tasks}.
\newblock In \emph{Proceedings of the 2019 Conference on Empirical Methods in
  Natural Language Processing and the 9th International Joint Conference on
  Natural Language Processing (EMNLP-IJCNLP)}, pages 2733--2743, Hong Kong,
  China. Association for Computational Linguistics.

\bibitem[{Hupkes et~al.(2018)Hupkes, Veldhoen, and
  Zuidema}]{hupkes2018visualisation}
Dieuwke Hupkes, Sara Veldhoen, and Willem Zuidema. 2018.
\newblock Visualisation and'diagnostic classifiers' reveal how recurrent and
  recursive neural networks process hierarchical structure.
\newblock \emph{Journal of Artificial Intelligence Research}, 61:907--926.

\bibitem[{Lee et~al.(2020)Lee, Tran, Firat, and
  Cho}]{lee-etal-2020-discrepancy}
Jason Lee, Dustin Tran, Orhan Firat, and Kyunghyun Cho. 2020.
\newblock \href {https://aclanthology.org/2020.spnlp-1.10} {On the discrepancy
  between density estimation and sequence generation}.
\newblock In \emph{Proceedings of the Fourth Workshop on Structured Prediction
  for NLP}, pages 84--94, Online. Association for Computational Linguistics.

\bibitem[{Li et~al.(2020)Li, Liu, Li, Li, Huang, and
  Shi}]{li-etal-2020-evaluating}
Jierui Li, Lemao Liu, Huayang Li, Guanlin Li, Guoping Huang, and Shuming Shi.
  2020.
\newblock \href {https://doi.org/10.18653/v1/2020.acl-main.35} {Evaluating
  explanation methods for neural machine translation}.
\newblock In \emph{Proceedings of the 58th Annual Meeting of the Association
  for Computational Linguistics}, pages 365--375, Online. Association for
  Computational Linguistics.

\bibitem[{Li et~al.(2019)Li, Li, Liu, Meng, and Shi}]{li-etal-2019-word}
Xintong Li, Guanlin Li, Lemao Liu, Max Meng, and Shuming Shi. 2019.
\newblock \href {https://doi.org/10.18653/v1/P19-1124} {On the word alignment
  from neural machine translation}.
\newblock In \emph{Proceedings of the 57th Annual Meeting of the Association
  for Computational Linguistics}, pages 1293--1303, Florence, Italy.
  Association for Computational Linguistics.

\bibitem[{Li et~al.(2018)Li, Liu, Tu, Shi, and Meng}]{li2018target}
Xintong Li, Lemao Liu, Zhaopeng Tu, Shuming Shi, and Max Meng. 2018.
\newblock \href {https://doi.org/10.18653/v1/N18-1125} {Target foresight based
  attention for neural machine translation}.
\newblock In \emph{Proceedings of the 2018 Conference of the North {A}merican
  Chapter of the Association for Computational Linguistics: Human Language
  Technologies}, pages 1380--1390, New Orleans, Louisiana. Association for
  Computational Linguistics.

\bibitem[{Long et~al.(2018)Long, Cao, Wang, and Jordan}]{long2018conditional}
Mingsheng Long, Zhangjie Cao, Jianmin Wang, and Michael~I. Jordan. 2018.
\newblock Conditional adversarial domain adaptation.
\newblock In \emph{Advances in Neural Information Processing Systems 31: Annual
  Conference on Neural Information Processing Systems 2018, NeurIPS 2018,
  December 3-8, 2018, Montr{\'{e}}al, Canada}, pages 1647--1657.

\bibitem[{Luong et~al.(2016)Luong, Le, Sutskever, Vinyals, and
  Kaiser}]{DBLP:journals/corr/LuongLSVK15}
Minh{-}Thang Luong, Quoc~V. Le, Ilya Sutskever, Oriol Vinyals, and Lukasz
  Kaiser. 2016.
\newblock Multi-task sequence to sequence learning.
\newblock In \emph{4th International Conference on Learning Representations,
  {ICLR} 2016, San Juan, Puerto Rico, May 2-4, 2016, Conference Track
  Proceedings}.

\bibitem[{Luong et~al.(2015)Luong, Pham, and Manning}]{luong2015effective}
Thang Luong, Hieu Pham, and Christopher~D. Manning. 2015.
\newblock \href {https://doi.org/10.18653/v1/D15-1166} {Effective approaches to
  attention-based neural machine translation}.
\newblock In \emph{Proceedings of the 2015 Conference on Empirical Methods in
  Natural Language Processing}, pages 1412--1421, Lisbon, Portugal. Association
  for Computational Linguistics.

\bibitem[{Mart{\'{\i}}nez et~al.(2020)Mart{\'{\i}}nez, Bertr{\'{a}}n, and
  Sapiro}]{martinez2020minimax}
Natalia Mart{\'{\i}}nez, Mart{\'{\i}}n Bertr{\'{a}}n, and Guillermo Sapiro.
  2020.
\newblock Minimax pareto fairness: {A} multi objective perspective.
\newblock In \emph{Proceedings of the 37th International Conference on Machine
  Learning, {ICML} 2020, 13-18 July 2020, Virtual Event}, volume 119 of
  \emph{Proceedings of Machine Learning Research}, pages 6755--6764. {PMLR}.

\bibitem[{Mas-Colell et~al.(1995)Mas-Colell, Whinston, Green
  et~al.}]{mas1995microeconomic}
Andreu Mas-Colell, Michael~Dennis Whinston, Jerry~R Green, et~al. 1995.
\newblock \emph{Microeconomic theory}, volume~1.
\newblock Oxford university press New York.

\bibitem[{McDonald and Chiang(2021)}]{mcdonald2021syntax}
Colin McDonald and David Chiang. 2021.
\newblock \href {https://doi.org/10.18653/v1/2021.naacl-srw.7} {Syntax-based
  attention masking for neural machine translation}.
\newblock In \emph{Proceedings of the 2021 Conference of the North American
  Chapter of the Association for Computational Linguistics: Student Research
  Workshop}, pages 47--52, Online. Association for Computational Linguistics.

\bibitem[{Merity et~al.(2018)Merity, Keskar, and
  Socher}]{merity2017regularizing}
Stephen Merity, Nitish~Shirish Keskar, and Richard Socher. 2018.
\newblock Regularizing and optimizing {LSTM} language models.
\newblock In \emph{6th International Conference on Learning Representations,
  {ICLR} 2018, Vancouver, BC, Canada, April 30 - May 3, 2018, Conference Track
  Proceedings}. OpenReview.net.

\bibitem[{N{\u{a}}dejde et~al.(2017)N{\u{a}}dejde, Reddy, Sennrich, Dwojak,
  Junczys-Dowmunt, Koehn, and Birch}]{DBLP:conf/wmt/NadejdeRSDJKB17}
Maria N{\u{a}}dejde, Siva Reddy, Rico Sennrich, Tomasz Dwojak, Marcin
  Junczys-Dowmunt, Philipp Koehn, and Alexandra Birch. 2017.
\newblock \href {https://doi.org/10.18653/v1/W17-4707} {Predicting target
  language {CCG} supertags improves neural machine translation}.
\newblock In \emph{Proceedings of the Second Conference on Machine
  Translation}, pages 68--79, Copenhagen, Denmark. Association for
  Computational Linguistics.

\bibitem[{Ott et~al.(2019)Ott, Edunov, Baevski, Fan, Gross, Ng, Grangier, and
  Auli}]{ott2019fairseq}
Myle Ott, Sergey Edunov, Alexei Baevski, Angela Fan, Sam Gross, Nathan Ng,
  David Grangier, and Michael Auli. 2019.
\newblock \href {https://doi.org/10.18653/v1/N19-4009} {fairseq: A fast,
  extensible toolkit for sequence modeling}.
\newblock In \emph{Proceedings of the 2019 Conference of the North {A}merican
  Chapter of the Association for Computational Linguistics (Demonstrations)},
  pages 48--53, Minneapolis, Minnesota. Association for Computational
  Linguistics.

\bibitem[{Papineni et~al.(2002)Papineni, Roukos, Ward, and
  Zhu}]{papineni2002bleu}
Kishore Papineni, Salim Roukos, Todd Ward, and Wei-Jing Zhu. 2002.
\newblock \href {https://doi.org/10.3115/1073083.1073135} {{B}leu: a method for
  automatic evaluation of machine translation}.
\newblock In \emph{Proceedings of the 40th Annual Meeting of the Association
  for Computational Linguistics}, pages 311--318, Philadelphia, Pennsylvania,
  USA. Association for Computational Linguistics.

\bibitem[{Pimentel et~al.(2020{\natexlab{a}})Pimentel, Saphra, Williams, and
  Cotterell}]{pimentel-etal-2020-Pareto}
Tiago Pimentel, Naomi Saphra, Adina Williams, and Ryan Cotterell.
  2020{\natexlab{a}}.
\newblock \href {https://doi.org/10.18653/v1/2020.emnlp-main.254} {{P}areto
  probing: {T}rading off accuracy for complexity}.
\newblock In \emph{Proceedings of the 2020 Conference on Empirical Methods in
  Natural Language Processing}, pages 3138--3153, Online. Association for
  Computational Linguistics.

\bibitem[{Pimentel et~al.(2020{\natexlab{b}})Pimentel, Valvoda, Hall~Maudslay,
  Zmigrod, Williams, and Cotterell}]{pimentel2020information}
Tiago Pimentel, Josef Valvoda, Rowan Hall~Maudslay, Ran Zmigrod, Adina
  Williams, and Ryan Cotterell. 2020{\natexlab{b}}.
\newblock \href {https://doi.org/10.18653/v1/2020.acl-main.420}
  {Information-theoretic probing for linguistic structure}.
\newblock In \emph{Proceedings of the 58th Annual Meeting of the Association
  for Computational Linguistics}, pages 4609--4622, Online. Association for
  Computational Linguistics.

\bibitem[{Qi et~al.(2020)Qi, Zhang, Zhang, Bolton, and Manning}]{qi2020stanza}
Peng Qi, Yuhao Zhang, Yuhui Zhang, Jason Bolton, and Christopher~D. Manning.
  2020.
\newblock \href {https://doi.org/10.18653/v1/2020.acl-demos.14} {{S}tanza: A
  python natural language processing toolkit for many human languages}.
\newblock In \emph{Proceedings of the 58th Annual Meeting of the Association
  for Computational Linguistics: System Demonstrations}, pages 101--108,
  Online. Association for Computational Linguistics.

\bibitem[{Ravfogel et~al.(2020)Ravfogel, Elazar, Gonen, Twiton, and
  Goldberg}]{ravfogel2020null}
Shauli Ravfogel, Yanai Elazar, Hila Gonen, Michael Twiton, and Yoav Goldberg.
  2020.
\newblock \href {https://doi.org/10.18653/v1/2020.acl-main.647} {Null it out:
  Guarding protected attributes by iterative nullspace projection}.
\newblock In \emph{Proceedings of the 58th Annual Meeting of the Association
  for Computational Linguistics}, pages 7237--7256, Online. Association for
  Computational Linguistics.

\bibitem[{Ravichander et~al.(2021)Ravichander, Belinkov, and
  Hovy}]{ravichander-etal-2021-probing}
Abhilasha Ravichander, Yonatan Belinkov, and Eduard Hovy. 2021.
\newblock \href {https://aclanthology.org/2021.eacl-main.295} {Probing the
  probing paradigm: Does probing accuracy entail task relevance?}
\newblock In \emph{Proceedings of the 16th Conference of the European Chapter
  of the Association for Computational Linguistics}, pages 3363--3377, Online.
  Association for Computational Linguistics.

\bibitem[{Saleh et~al.(2020)Saleh, Deutsch, Casper, Belinkov, and
  Shieber}]{saleh2020probing}
Abdelrhman Saleh, Tovly Deutsch, Stephen Casper, Yonatan Belinkov, and Stuart
  Shieber. 2020.
\newblock \href {https://doi.org/10.18653/v1/2020.nlp4convai-1.15} {Probing
  neural dialog models for conversational understanding}.
\newblock In \emph{Proceedings of the 2nd Workshop on Natural Language
  Processing for Conversational AI}, pages 132--143, Online. Association for
  Computational Linguistics.

\bibitem[{Sennrich and Haddow(2016)}]{sennrich2016linguistic}
Rico Sennrich and Barry Haddow. 2016.
\newblock \href {https://doi.org/10.18653/v1/W16-2209} {Linguistic input
  features improve neural machine translation}.
\newblock In \emph{Proceedings of the First Conference on Machine Translation},
  pages 83--91, Berlin, Germany. Association for Computational Linguistics.

\bibitem[{Sennrich et~al.(2016)Sennrich, Haddow, and
  Birch}]{sennrich2015neural}
Rico Sennrich, Barry Haddow, and Alexandra Birch. 2016.
\newblock \href {https://doi.org/10.18653/v1/P16-1162} {Neural machine
  translation of rare words with subword units}.
\newblock In \emph{Proceedings of the 54th Annual Meeting of the Association
  for Computational Linguistics}, pages 1715--1725, Berlin, Germany.
  Association for Computational Linguistics.

\bibitem[{Sutskever et~al.(2014)Sutskever, Vinyals, and
  Le}]{sutskever2014sequence}
Ilya Sutskever, Oriol Vinyals, and Quoc~V. Le. 2014.
\newblock Sequence to sequence learning with neural networks.
\newblock In \emph{Advances in Neural Information Processing Systems 27: Annual
  Conference on Neural Information Processing Systems 2014, December 8-13 2014,
  Montreal, Quebec, Canada}, pages 3104--3112.

\bibitem[{Tzeng et~al.(2017)Tzeng, Hoffman, Saenko, and
  Darrell}]{tzeng2017adversarial}
Eric Tzeng, Judy Hoffman, Kate Saenko, and Trevor Darrell. 2017.
\newblock \href {https://doi.org/10.1109/CVPR.2017.316} {Adversarial
  discriminative domain adaptation}.
\newblock In \emph{2017 {IEEE} Conference on Computer Vision and Pattern
  Recognition, {CVPR} 2017, Honolulu, HI, USA, July 21-26, 2017}, pages
  2962--2971. {IEEE} Computer Society.

\bibitem[{Vapnik(1999)}]{vapnik1999overview}
Vladimir~N Vapnik. 1999.
\newblock An overview of statistical learning theory.
\newblock \emph{IEEE transactions on neural networks}, 10(5):988--999.

\bibitem[{Vaswani et~al.(2017)Vaswani, Shazeer, Parmar, Uszkoreit, Jones,
  Gomez, Kaiser, and Polosukhin}]{vaswani2017attention}
Ashish Vaswani, Noam Shazeer, Niki Parmar, Jakob Uszkoreit, Llion Jones,
  Aidan~N. Gomez, Lukasz Kaiser, and Illia Polosukhin. 2017.
\newblock Attention is all you need.
\newblock In \emph{Advances in Neural Information Processing Systems 30: Annual
  Conference on Neural Information Processing Systems 2017, December 4-9, 2017,
  Long Beach, CA, {USA}}, pages 5998--6008.

\bibitem[{Voita and Titov(2020)}]{voita2020information}
Elena Voita and Ivan Titov. 2020.
\newblock \href {https://doi.org/10.18653/v1/2020.emnlp-main.14}
  {Information-theoretic probing with minimum description length}.
\newblock In \emph{Proceedings of the 2020 Conference on Empirical Methods in
  Natural Language Processing (EMNLP)}, pages 183--196, Online. Association for
  Computational Linguistics.

\bibitem[{Xie et~al.(2017)Xie, Dai, Du, Hovy, and Neubig}]{xie2017controllable}
Qizhe Xie, Zihang Dai, Yulun Du, Eduard~H. Hovy, and Graham Neubig. 2017.
\newblock Controllable invariance through adversarial feature learning.
\newblock In \emph{Advances in Neural Information Processing Systems 30: Annual
  Conference on Neural Information Processing Systems 2017, December 4-9, 2017,
  Long Beach, CA, {USA}}, pages 585--596.

\bibitem[{Xu et~al.(2020)Xu, Li, Cui, Huang, Wei, and Zhou}]{xu2020layoutlm}
Yiheng Xu, Minghao Li, Lei Cui, Shaohan Huang, Furu Wei, and Ming Zhou. 2020.
\newblock Layoutlm: Pre-training of text and layout for document image
  understanding.
\newblock In \emph{{KDD} '20: The 26th {ACM} {SIGKDD} Conference on Knowledge
  Discovery and Data Mining, Virtual Event, CA, USA, August 23-27, 2020}, pages
  1192--1200. {ACM}.

\bibitem[{Zaremba et~al.(2014)Zaremba, Sutskever, and
  Vinyals}]{zaremba2014recurrent}
Wojciech Zaremba, Ilya Sutskever, and Oriol Vinyals. 2014.
\newblock Recurrent neural network regularization.
\newblock \emph{arXiv preprint arXiv:1409.2329}.

\end{thebibliography}
\bibliographystyle{acl_natbib}

\clearpage

\appendix

\section{Training Details\label{appendix:train}}
On the WMT14 corpus, training one LSTM model with 4 V100 GPUs costs 5 hours, and training one Transformer with 8 V100 GPUs costs 8 hours. On LDC corpus, training one LSTM model with 4 V100 GPUs costs 3 hours, and training one Transformer with 8 V100 GPUs costs 3 hours. On the PTB dataset, training LSTM model with 1 V100 GPU costs 6 minutes.

When running our algorithm, we empirically observe that when $\lambda$ is below 0.01, the optimized models show little difference comparing with the standard model, and when $\lambda$ is larger than 0.1, the proposed algorithm becomes unstable and can't converge to Pareto-optimal solutions well. Therefore, we take ten values from 0.1 to 0.01 at equal intervals as $\lambda$ in Eq.~\ref{eq:interpolating}, and train ten models with different $\lambda$ for each condition respectively. Then we plot all the models and the Pareto frontier of these models in the experiments. 

\section{Effects of Randomness\label{sec:rand}}
\begin{table}[]
\centering
\resizebox{0.36\textwidth}{!}{
\begin{tabular}{cccc}
\toprule
\multicolumn{2}{c}{BLEU} & \multicolumn{2}{c}{H(POS|h)} \\ \cmidrule(r){1-2} \cmidrule(r){3-4}
mean       & var         & mean           & var        \\ \midrule
21.08      & 0.00407     & 0.1113         & 0          \\
21.32      & 0.01536     & 0.1093         & 0          \\
21.49      & 0.01847     & 0.108          & 0          \\
21.52      & 0.00060     & 0.1123         & 0          \\ \bottomrule
\end{tabular}}
\caption{Experiment results from LSTM + POS setting. Specifically, ``mean'' and ``var'' denotes the mean and the variance over the window.}
\label{randomness}
\end{table}

Following the method from \citet{chen-etal-2018-best}, we check if randomness will affect our experimental results. Specifically, we select a window of size 3 around the best checkpoint model and report the mean and variance over the selected window. The results are shown in Table~\ref{randomness}. Because repeating experiments under all the settings are too extensive, we only randomly select 4 models from LSTM + POS settings. As shown in the table, all the variances are small, and the variances of the entropy even achieve 0. This suggests that the random disturbance of our experiments are small and thus our results are reliable.


\end{document}